\documentclass[11pt,a4paper]{article}
\usepackage[hyperref]{acl2020}
\usepackage{times}
\usepackage{latexsym}

\usepackage{microtype}
\usepackage{graphicx}

\usepackage{times}
\usepackage{algorithm}
\usepackage{algorithmic}
\usepackage{IEEEtrantools}
\usepackage{booktabs}
\usepackage{amsfonts}

\newcommand{\tabincell}[2]{\begin{tabular}{@{}#1@{}}#2\end{tabular}}

\aclfinalcopy 

\title{Pre-training Text Representations as Meta Learning}

\author{Shangwen Lv\textsuperscript{\rm 1}\thanks{\ Euqal Contributions. Work was done while this author was an intern at Microsoft Research Asia.} , Yuechen Wang\textsuperscript{\rm 2}\footnotemark[1] , Daya Guo\textsuperscript{\rm 3}\footnotemark[1] ,  Duyu Tang\textsuperscript{\rm 4}, Nan Duan\textsuperscript{\rm 4}, Fuqing Zhu\textsuperscript{\rm 1}, \\  
\textbf{Ming Gong\textsuperscript{\rm 4}, Linjun Shou\textsuperscript{\rm 4}, Ryan Ma\textsuperscript{\rm 4}, Daxin Jiang\textsuperscript{\rm 4}, Guihong Cao\textsuperscript{\rm 4}, Ming Zhou\textsuperscript{\rm 4}, Songlin Hu\textsuperscript{\rm 1}}\\
\textsuperscript{\rm 1}Institute of Information Engineering, Chinese Academy of Sciences\\
\textsuperscript{\rm 2} University of Science and Technology of China \
\textsuperscript{\rm 3}Sun Yat-sen University \
\textsuperscript{\rm 4} Microsoft Corporation \\
\{lvshangwen, zhufuqing, husonglin\}@iie.ac.cn \\
wyc9725@mail.ustc.edu.cn, guody5@mail2.sysu.edu.cn,
\\
\{dutang,nanduan,migon,lisho,ryanma,djiang,gucao,mingzhou\}@microsoft.com \\
}

\date{}

\begin{document}
\maketitle
\begin{abstract}
Pre-training text representations has recently been shown to significantly improve the state-of-the-art in many natural language processing tasks.
The central goal of pre-training is to learn text representations that are useful for subsequent tasks. However, existing approaches are optimized by minimizing a proxy objective, such as the negative log likelihood of language modeling.
In this work, 
we introduce a learning algorithm which directly optimizes model's ability to learn text representations for effective learning of downstream tasks.
We show that there is an intrinsic connection between multi-task pre-training and model-agnostic meta-learning with a sequence of meta-train steps.
The standard multi-task learning objective adopted in BERT is a special case of our learning algorithm where the depth of meta-train is zero. 
We study the problem in two settings: unsupervised pre-training and supervised pre-training with different pre-training objects to verify the generality of our approach.
Experimental results show that our algorithm brings improvements and learns better initializations for a variety of downstream tasks.
\end{abstract}



\section{Introduction}

\label{section:intro}
\begin{figure}[t]
	\centering
	\includegraphics[width=0.48\textwidth]{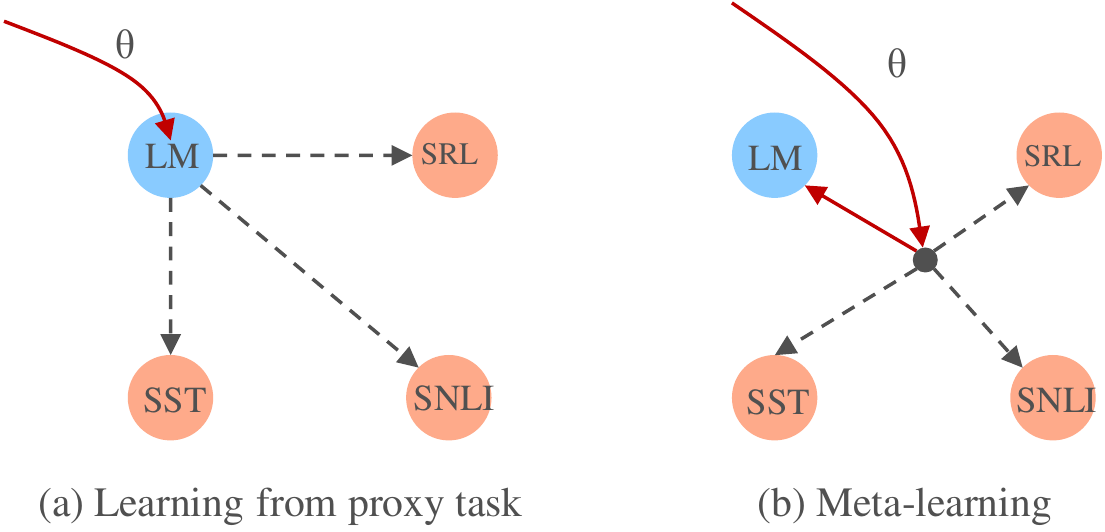}
	\caption {An illustration of pre-training as (a) a proxy task like language modeling and (b) meta-learning. Solid lines in red present pre-training. Dashed lines represent task-specific fine-tuning.}
	\label{fig:intro}
\end{figure}

The primary goal of pre-training text representations is to acquire useful representations from data that can be effectively used for learning downstream NLP tasks.
Although pre-trained models bring significant gains in many NLP tasks recently\cite{rajpurkar2016squad,zellers2018swag},  these approaches are learned by optimizing a proxy task, such as language modeling \cite{peters18-elmo,howard18,radford2018improving,devlin2018bert}, machine translation \cite{mccann2017learned}, next sentence generation \cite{kiros2015skip}, discourse coherence \cite{jernite2017discourse}, etc. 
These objectives are different from the primary goal of pre-training, and result in the mismatch between the pre-training and fine-tuning.
An illustrative example is given in Figure \ref{fig:intro}.

This paper explores to alleviate the mismatch between pre-training and fine-tuning processes. 
Pre-training does not have an explicit learning objective like a standard optimization problem, yet it fits well to the meta objective in  meta learning \cite{schmidhuber1987evolutionary,bengio1992optimization}, which is to train a good learner measured by the learner's performance on downstream (maybe unseen) tasks.  The learning process is akin to how humans build upon their prior experience and use them to quickly learn new concepts. 

We present a learning algorithm to directly optimize model's ability to learn a representation of text for its application on downstream tasks.
We show that there is an intrinsic connection between the multi-task objectives for pre-training and Model-Agnostic Meta-Learning (MAML) \cite{finn2017model} with a sequence of meta-train steps. 
When the number of meta-train step is zero, the learning algorithm falls back to the standard multi-task learning objective used in BERT \cite{devlin2018bert}.


We perform experiments on unsupervised pre-training and supervised pre-training with different pre-training objects to verify the generality of our approach. We conduct experiments on a more light-weighted pre-trained model ELMo~\cite{peters18-elmo} to evaluate the ability of our approach to learn from scratch.  
Comprehensive experiments show that our pre-trained model outperforms BERT on diverse downstream tasks. Meanwhile, our learning algorithm can also learn a better initialization than BERT on various downstream tasks. To the best of our knowledge, this is the first work that explores meta learning for pre-training text representations.


\section{Related Work}
\paragraph{Pre-training Text Representation} Pre-trained text representations from unlabeled corpora have proven effective for many NLP tasks.
Earlier works focus on learning embeddings for \mbox{words} \cite{mikolov2013distributed,penningtonSM14}, the basic idea of which is to represent a word with \mbox{its} surrounding contexts.
Recent studies show that pre-trained embeddings for longer pieces of \mbox{text} (e.g. a sentence, paraphrase, document) and
contextualized word embeddings \cite{peters18-elmo,howard18,radford2018improving,devlin2018bert,liu2019mt-dnn,Dong2019UnifiedLM}  are surprisingly useful, even drive the state-of-the-art to achieve human-level accuracy on challenging datasets like SQuAD \cite{rajpurkar2016squad} and SWAG \cite{zellers2018swag}.
Existing works in this direction typically optimize the pre-trained model using a certain task such as language modeling.
However, a natural question is why should we learn representations optimized by language modeling?
The goal of pre-training text representation is not language modeling, but learning useful representations for downstream tasks.
In this work, we directly optimize the pre-trained model towards this goal and leverage successful meta-learning algorithm MAML.

\paragraph{Meta-Learning}
Meta-learning,
or learning to learn, is a promising direction to deal with few-shot learning with the ability to quickly learn for new tasks by reusing previous experience.  
We briefly summarize existing approaches into three categories.
The {first} category aims to learn a representation.
The idea is to learn a useful representation for each example, such that examples from the same category are close while examples from different categories are far apart.
Matching networks \cite{vinyals2016matching} measure the similarity at the datapoint-level. 
Prototypical networks \cite{snell2017} consider representation at the task-level by aggregating representations of the examples for each category.
The {second} category aims at learning an optimizer.
\citet{ravi2017optimization} uses an LSTM as the meta-learner to learn to update the learner, so that the learner quickly learns for a new task.
\citet{finn2017model} introduces Model-Agnostic Meta-Learning (MAML), which is optimized for a good initial representation that can be quickly fine-tuned from examples in a new task. We follow \cite{finn2017model} and use MAML in this paper.
The {third} category aims at learning a \mbox{recurrent} \cite{santoro2016meta} or temporal convolutional \cite{mishra2017meta} neural network that uses previous experience.
Meta-learning has been used for low-resource neural machine translation \cite{gu18meta} and semantic parsing \cite{huang18}. To the best of our knowledge, this is the first work that explores meta-learning for pre-training text representations.

\section{Approach}
We first analyze the multi-task learning objective for pre-training \mbox{text} representations and its connection with model-agnostic meta-learning. 
And then we present a computationally efficient learning algorithm based on  approximation strategies. 



\subsection{Multi-task Pre-training as Meta Learning}
\label{section:pre-training-as-maml}
We consider the problem as learning a mapping function $f(s) \rightarrow \mathbb{R}^d$ parameterized by $\theta$, which maps any text $s$ to a continuous vector whose dimension is $d$. 
Let $M_{\theta_0}$ be a pre-trained model with parameters $\theta_0$ which is to be learned.
Take a certain downstream task $\mathcal{T}_i$ as an example. Let's denote its training data as $D_{\mathcal{T}_i}^{train}$, its evaluation data as $D_{\mathcal{T}_i}^{test}$, and it loss function as $\mathcal{L}_{\mathcal{T}_i}(\theta; D)$.

The objective of pre-training is maximizing the performance on various downstream tasks, equivalent to minimizing the loss function of the fine-tuned parameter $\theta_k$ over the test data $D_{\mathcal{T}_i}^{test}$. The fine-tuned parameter $\theta_k$ is calculated with multiple (e.g. $k$) gradient descent steps over the training data $D_{\mathcal{T}_i}^{train}$, staring from the pre-trained parameter $\theta_0$.
Equation \ref{gradient_descent} shows the calculation process, where $D_{\mathcal{T}_i}^{train_j}$ is the $j$-th batch of training examples and $\alpha$ is the learning rate of the fine-tuning process.
\begin{IEEEeqnarray}{rl}
\label{gradient_descent}
\theta_{k}&=\theta_{k-1}-\alpha\nabla_{\theta_{k-1}}\mathcal{L}_{\mathcal{T}_i}(\theta_{k-1}; D_{\mathcal{T}_i}^{train_k}) \,, \nonumber\\ 
\qquad ... &    \nonumber \\
\theta_{2}&=\theta_{1}-\alpha\nabla_{\theta_{1}}\mathcal{L}_{\mathcal{T}_i}(\theta_{1}; D_{\mathcal{T}_i}^{train_2}) \,, \nonumber \\
\theta_1&=\theta_{0}-\alpha\nabla_{\theta_{0}}\mathcal{L}_{\mathcal{T}_i}(\theta_{0}; D_{\mathcal{T}_i}^{train_1}) \,.
\end{IEEEeqnarray}

We denote $\theta_k=f_k (\theta_0)$, our pre-training object then becomes:
\begin{IEEEeqnarray}{rl}
\label{objective}
\theta_0 &=\mathop{\arg\min}_{\theta_0} \mathcal{L}_{\mathcal{T}_i}(\theta_k; D_{\mathcal{T}_i}^{test}) \nonumber \\
&= \mathop{\arg\min}_{\theta_0} \mathcal{L}_{\mathcal{T}_i}(f_k (\theta_0); D_{\mathcal{T}_i}^{test})\,.
\end{IEEEeqnarray}

Equation \ref{objective} means that our pre-training object is to find a optimal $\theta_0$ to minimize the fine-tuning loss on test dataset.

Let $\mathcal{T}_{p}$ denote the multi-task pre-training tasks.  Our pre-training procedure should include procedures similar to fine-tuning on training dataset and evaluate the fine-tuned model on the test dataset. 
We first fetch $k$ batch pre-training data $D_{\mathcal{T}_{p}}^{train}$ and perform a series of train steps similar to Equation \ref{gradient_descent} to get $\theta_k '=f_k(\theta_0 ')$ where $\theta_0 '$ is the initialized parameters of pre-trained models. In order to mimic the fine-tuning evaluation on test dataset, we fetch one batch of pre-training data as the test batch $D_{\mathcal{T}_{p}}^{test}$ since we cannot foreseen or assume downstream tasks. And then perform evaluation on the test batch. Finally, we update $\theta_0 '$ as follows:

\begin{equation}\label{update}
\theta_0 '=\theta_0 '-\beta\nabla_{\theta_0 '}\mathcal{L}_{\mathcal{T}_{p}}(\theta_k '; D_{\mathcal{T}_{p}}^{test}) \,,
\end{equation}
where $\beta$ is the learning rate of the training process.

Following the chain rule, we can rewrite the gradient $\nabla_{\theta_0 '}\mathcal{L}_{\mathcal{T}_{p}}(\theta_k ')$ as follows:


\begin{small}
\begin{IEEEeqnarray}{ll}
\label{chain-rule}
&\nabla_{\theta_0 '}\mathcal{L}_{\mathcal{T}_{p}}(\theta_k '; D_{\mathcal{T}_{p}}^{test}) \\
&=\nabla_{\theta_k '}\mathcal{L}_{\mathcal{T}_{p}}  (\theta_k '; D_{\mathcal{T}_{p}}^{test}) \times(\nabla_{\theta_{k-1} '}\theta_k ')\cdots \times (\nabla_{\theta_{0} '}\theta_1 ') \nonumber
\\&=\nabla_{\theta_k '}\mathcal{L}_{\mathcal{T}_{p}}(\theta_k '; D_{\mathcal{T}_{p}}^{test})\prod_{j=1}^{k}(I-\nabla^{(2)}_{\theta_{j-1} '}\mathcal{L}_{\mathcal{T}_{p}}(\theta_{j-1} '; D_{\mathcal{T}_{p}}^{train_j})) \,, \nonumber
\end{IEEEeqnarray}
\end{small}
where $D_{\mathcal{T}_{p}}^{train_j}$ is the $j$ batch of data in $D_{\mathcal{T}_p}^{train}$.

Equation \ref{chain-rule} aligns well to MAML. The learning process of MAML includes two steps, a meta-train process which quickly updates the model parameter using gradient descent over a meta-train set, and a meta-test process which measures the goodness of the updated/new parameter on a meta-test set. For consistency, we denote the pre-training step on $D_{\mathcal{T}_p}^{train}$ and $D_{\mathcal{T}_p}^{test}$ as meta train steps and meta test steps respectively.


The learning procedure of BERT is an oversimplified example of our pre-training procedure with meta train step $k$=0. Our learning algorithm is summarized in Algorithm~\ref{alg:example}. The learned $\theta_0'$ is the obtained pre-training parameter.




\begin{algorithm}[h]
 	\caption{Pre-training Text Representations as Meta Learning}
	\label{alg:example}
	\begin{algorithmic}[1]
		\REQUIRE  $p(\mathcal{T})$: distribution of pre-training tasks 
		\REQUIRE  $\alpha$, $\beta$: step size hyper-parameters
		\STATE Initialize $\theta_0 '$
			\WHILE{not done}
			\STATE Sample $k$ batches of data $\mathcal{D}_{\mathcal{T}_p}^{train}$ from multiple tasks following $p(\mathcal{T})$
			\FOR{$j$ from 1 to $k$}
				 \STATE Compute gradients using $\mathcal{L}(\theta_{j-1} '; \mathcal{D}_{\mathcal{T}_p}^{train_j}$) 
				 \STATE Update $\theta_j '$ based on Equation \ref{gradient_descent}
			\ENDFOR
			\STATE Sample a batch of data $\mathcal{D}_{\mathcal{T}_p}^{test}$ following  $p(\mathcal{T})$ \\
			\STATE Compute gradients using $\mathcal{L}(\theta_k '; \mathcal{D}_{\mathcal{T}_p}^{test})$ 
			\STATE Update $\theta_0 '$ based on Equation \ref{update} and \ref{chain-rule}
			
		\ENDWHILE
	\end{algorithmic}
\end{algorithm}

\subsection{Efficient Implementation}
\label{section:learning-algo}
In practice, calculating derivatives at high-order is expensive.
As suggested by Finn et al. \cite{finn2018learning}, first-order approximation can save around 33\% of the computation, while achieving similar performance to including full second-order information on few-shot image recognition benchmarks. 
The approximated update rule is given as follows:
\begin{equation}
\theta_0 '\approx\theta_0 '-\beta\nabla_{\theta_k '}\mathcal{L}_{\mathcal{T}_p}(\theta_k '; D_{\mathcal{T}_p}^{test}) \,,
\end{equation}
where the second-order information in Equation 4 is ignored.

\section{Experiment}
In this section, we conduct experiments on unsupervised tasks pre-training and supervised tasks pre-training to show the generality of our approach. Furthermore, we verify the ability of our algorithm to pre-train from scratch with a more light-weighted  model ELMo~\cite{peters18-elmo}.
Experiments show that our pre-training method can achieve better results and learn better initializations for downstream tasks. 

\subsection{Unsupervised Tasks as Pre-training Tasks}


For unsupervised pre-training tasks, we utilize the same two tasks masked language model and next sentence prediction, following  the  pre-training  multi-tasks  in  BERT~\cite{devlin2018bert}.  WordPiece \cite{wu2016google} is adopted to  split words into tokens and we denote the split word pieces with \#\#. The maximum length of input sequence is 512. The two tasks are illustrated as follows:

\textbf{Masked Language Model} The Masked Language Model is also known as Cloze Task \cite{taylor1953cloze}. Some tokens are masked and the model targets at predicting the masked tokens. Following BERT \cite{devlin2018bert}, we randomly masked 15\% of the tokens of all WordPice tokens in the sentences.  Among masked positions, we replace the masked position token with \texttt{[MASK]} 80\% of the time. 10\% of the time we replace the masked position tokens with a randomly chosen token, and 10\% of the time we keep the original token. 

\textbf{Next Sentence Prediction}  The Next Sentence Prediction task aims to understand the \textit{relationship} between sentences \texttt{A} and \texttt{B} such as Question Answering tasks and Pair-wise Sentence Matching tasks. Following BERT \cite{devlin2018bert}, we choose the sentence \texttt{A} and \texttt{B} for each training example as follow: 50\% of the time, \texttt{B} is the actual next sentence after \texttt{A}, while 50\% of the time \texttt{B} is randomly chosen from the corpus.

For the pre-training corpus, we adopt the concatenation of English Wikipedia{\footnote{Wikipedia version enwiki-20190301}} and BookCorpus\footnote{https://yknzhu.wixsite.com/mbweb}. We only use the text passages in Wikipedia and ignore the tables, lists and headers\footnote{We adopt Wikipedia processing tool at https://github.com/attardi/wikiextractor}. 

We follow the similar model size and pretraining settings as BERT-base \cite{devlin2018bert}. Specifically, we use a 12-layer Transformer with 768 hidden size and 12 attention heads, which contains about 110M parameters. The model parameters are initialized with official BERT-base \cite{devlin2018bert}. The gelu activation \cite{hendrycks2016bridging} is used as in BERT \cite{devlin2018bert}.

Our implementation is based on the PyTorch implementation of BERT\footnote{https://github.com/huggingface/pytorch-pretrained-BERT}. We initialize our model with official BERT-base parameters. We use Adam \cite{kingma2014adam} with $\beta_1$ = 0.9, $\beta_2$ = 0.999 for optimization. The learning rate is set to 2e-5. The dropout rate is 0.1 and the weight decay is 0.01. The batch size is set to 128 to fully utilize the GPU memories. We run the pre-training procedure for about $240,000$ meta test steps. For pre-training with $k$ meta train steps, the total pre-training steps will be $(k+1)*240,000$ steps. It takes about 5 hours for 68,000 steps using 8 Nvidia Telsa V100 16GB GPU cards with mixed precision training.

We select different meta train steps to verify the effectiveness of our approach. Specifically, we set meta train steps $k \in \{1,3,5,10,20\}$. For fair comparison of each setting, we pre-train for the same meta test steps with each setting.

We perform a variety of downstream tasks to verify the effectiveness of our pre-training procedure. We perform experiments on single sentence classification, pair-wise sentence matching and cloze tasks. For single sentence classification tasks, we adopt sentiment classification tasks SST-2 and SST-5. For pair-wise sentence matching, we adopt MNLI and SNLI. We adopt CLOTH as our testbed for cloze tasks. 

\textbf{SST-2} \ The Stanford Sentiment Treebank-2 \cite{socher2013recursive} is to classify the sentiment of one given sentence. Each sentence is classified into two categories: positive or negative. The sentences are retrieved from movie reviews and have human-annotated labels.

\textbf{SST-5} \ The Stanford Sentiment Treebank-5 \cite{socher2013recursive} is also a sentiment classification dataset. Different from SST-2 which only has two sentiment categories, SST-5 has five fine-grained sentiment categories, from very negative to very positive to describe a movie review.

\textbf{MNLI} Multi-Genre Natural Language Inference \cite{nangia2017repeval} is a large-scale entailment classification task. Each sentence pair has one hypothesis with a label \textit{entailment}, \textit{contradiction} or \textit{neutral} with respect to the premise. The development and test datasets are split into in-domain (matched) and out-domain (mismatched) datasets. 

\textbf{SNLI} The Stanford Natural Language Inference dataset \cite{bowman2015large} format is similar to that of MNLI. It consists of 570k human-annotated sentence pairs. The premises are derived from the Flickr30 corpus captions and the hypothesis are manually annotated.


\textbf{CLOTH} \textbf{CLO}ze test by \textbf{T}eac\textbf{H}ers dataset \cite{xie2017large} is collected from three free and public websites in China that gather exams created by English teachers to prepare students entrance exams. It contains high school and middle school exams. The task is to select the right answer from four candidate answers according to the context.

The first part in Table \ref{unsupervised_dataset} shows the dataset distributions and the accurcy is the metric to measure the performance of different models. 

\begin{table}[htbp]
    \centering
    \begin{tabular}{c|c|c|c|c}
    \toprule
    \textbf{Corpus}  & \textbf{\#Train} & \textbf{\#Dev}  & \textbf{\#Test}  & \textbf{\#Label}  \\
    \midrule
    SST-2 &    67k  &  872  &  1.8K  &   2      \\
    
    SST-5 &  8.5K  &  1.1K  & 2.2K   &  5  \\
   
    SNLI &  549k  &  9.8k  & 9.8k   &   3     \\
    
    MNLI &  393K &  20K  &  20K  &  3  \\
  
    CLOTH-M & 22K & 3.3K & 3.2K & 4\\

    CLOTH-H  &  54.8K &  7.8K & 8.3K  &    4    \\
    \midrule
    \midrule
    QTC  & 3,074  &   384  &  384  &  2    \\
    QDC  & 45,833  &  4,108  &  4,108  &    2    \\
    QPP&   4,234   &  760  &  760  &  2 \\
    \bottomrule
    \end{tabular}
    \caption{Statistics of fine-tuning datasets. The first part is unsupervised fine-tunintg datasets and the second part is the supervised fine-tuning datasets.}
    \label{unsupervised_dataset}
\end{table}

\subsection{Supervised Tasks as Pre-training Tasks}
For supervised pre-training tasks, we utilize the question-answer pair matching and question-question pair matching as the object of multi-task pre-training. Question-answer pair matching aims to determine if the given answer can answer the question properly and question-question pair matching aims to determine if two questions have the same meaning.

The settings with question-answer pair matching and question-question pair matching are similar to the settings in the former section. We set the input length to 128. During the pre-training, we adopt two tasks to perform pre-training:

\textbf{Question-Answer Pair Matching} The Question-Answer pair from search engines contains 4M human-labeled Question-Answer pairs. Each example contains a question, an answer and a label 1 or 0 denoting whether the given answer can answer the question or not. 

\textbf{Question-Question Pair Matching} The Question-Question pair from search engines contains 1M human-labeled Question-Question pairs. Each example contains two questions and one label 1 or 0 denoting whether the two questions are semantically equivalent or not.

The two tasks share the same BERT-base encoder, with two task-specific linear layers. The final loss is the sum of two task losses. We also initialize our model with official BERT-base parameters.

The batch size is set to 768 to fully utilize the GPU memories. We run the pre-training procedure for about $250,000$ meta test steps. For pre-training with $k$ meta train steps, the total training step will be $(k+1)*250,000$ steps. It takes about 5 hours for 330,000 steps using 8 Nvidia Telsa V100 16GB GPU cards with mixed precision training.


For the downstream tasks, we utilize three query-related tasks in search engines to verify the effectiveness of our pre-trained models, including QTC, QDC and QPP. These three tasks are utilized in search engines to provide evidence for performing knowledge base question answering.

\textbf{QTC} The Query-Type Classification task aims to predict if a query contains a single predict or not. Each instance contains a query and a label 0 or 1 indicating whether the query contains one single predicate or not. Accuracy is adopted to measure model performances.

\textbf{QDC} The Query-Domain Classification task aims to determine if the query belongs to a specific domain. Each example contains a query and a label indicating whether the query belongs to the specific domain or not. We adopt the movie domain as the testbed.  Accuracy is adopted as the evaluation metric.

\textbf{QPP} The Query-Predicate Pair datasets are extracted from search engines. Each instance contains one query and one predicate, with a label 1 or 0 indicating whether the query is equivalent to the predicate. Accuracy is also adopted as the evaluation metric.

The positive and negative instances are balanced in these three datasets. The detailed dataset statistics are shown in the second part in Table \ref{unsupervised_dataset}.






\subsection{Experiment Results}

The results are shown in Table \ref{unsupervised_result}. In the first group, we adopt masked language model and next sentence prediction as the pre-training tasks. In the second group, we adopt question-answer pair matching and question-question pair matching as the pre-training tasks. $k$=0 denotes we adopt the official BERT-base for fine-tuning, while $k\in\{1,3,5,10,20\}$ means we adopt the pre-trained model which performs $k$ meta train steps followed by one meta test step during pre-training. 

\begin{table*}[h]
    \centering
    \begin{tabular}{c|c|c|c|c|c|c}
    \toprule
    \textbf{Dataset}  &  \textbf{BERT-base ($k$=0)}&  \textbf{$k$=1}  &  \textbf{$k$=3}  & \textbf{$k$=5} & \textbf{$k$=10} & \textbf{$k$=20} \\
    \midrule
    SST-2 &  93.50 \% & 93.85\%  &  94.01\% & 93.79\%  &  \textbf{94.23\%} & 93.82\%  \\
    
    SST-5 & 54.84\% & 54.80\%  & 54.89\% & 55.71\%  & \textbf{55.97\%} & 54.98\% \\
   
    SNLI & 90.80\% & 90.80\% &  90.89\% &  \textbf{91.12\%}  & 91.10\% & 90.89\% \\
  
    MNLI matched &  84.60\% & 84.70\% & 84.70\% & \textbf{84.90\%} & 84.70\% & 84.65\% \\
    
    MNLI mismatched & 83.40\% &   83.50\%  &  83.60\%  &  83.40\%  &   \textbf{83.73\%} & 83.45\%\\
  
    CLOTH  & 82.00\%   &  82.22\%  & 82.27\%  &  82.22\% &  \textbf{82.40\%}  & 82.10\% \\
    
    CLOTH-M  & 85.00\%  &   85.37\%     & 85.46\%  &  \textbf{85.49\%}    & 85.37\% &  85.25\%\\
    CLOTH-H  & 80.90\%  &   81.01\%    & 81.04\%  &   80.97\%   & \textbf{81.25\%} & 81.05\% \\
    
    \midrule
    \midrule
    
    QTC  & 75.26\%  &  76.30\%  &  76.02\%  &  76.56\%  & \textbf{77.86\%}  & 76.82\% \\
    QDC  & 84.76\%  & 85.44\%   &  85.69\%  &  \textbf{86.05\%}  & 85.93\%  & 85.49\% \\
    QPP& 75.53 \%  &  76.71\%   &  76.58\%  &  \textbf{76.97\%}  &  76.18\%  & 75.66\%  \\ 
    
    \bottomrule
    \end{tabular}
    \caption{Fine-tuning results on diverse downstream tasks.}
    \label{unsupervised_result}
\end{table*}

\begin{figure*}[ht]
	\centering
	\includegraphics[width=\textwidth]{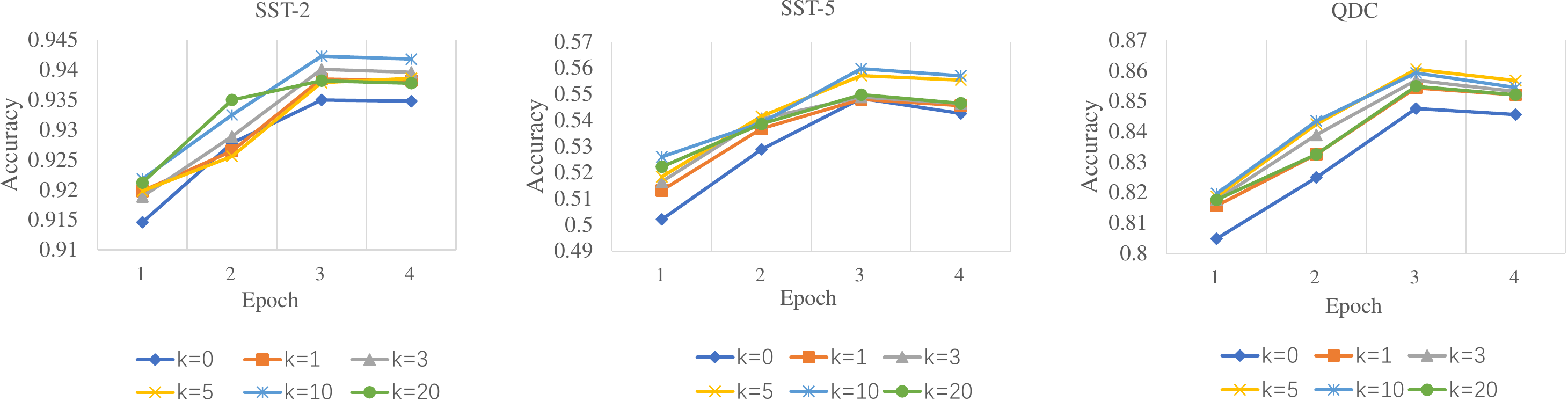}	
	\caption {Results on each epoch for SST-2, SST-5 and QDC datasets. The meta train step $k\geq 1$ has a better initialization than $k=0$ on various downstream tasks at epoch 1.}
	\label{fig:curve}
\end{figure*}
 

From the results, we observe that our training algorithm can outperform BERT-base on a variety of downstream tasks, including both the pre-training settings with different pre-training objectives.

 We also observe that when we increase the meta train step $k$ from 1 to 10, the results continue to increase and the best results are mostly got at $k$=5 or $k$=10. The results verify that the learned text representations with meta learning approach is more 
 beneficial for fine-tuning on downstream tasks.
 However, we observe that when we increase $k$ to 20, the results start to drop. When meta train step is large, it may cause the  gradients deviate much from the normal ones and will not provide enough information for the learning process.



\subsection{Pre-training from Scratch}
In this part, we test the ability of our algorithm to pre-train from scratch. It is heavy to pre-train BERT from scratch, so we select biLSTM-based ELMo~\cite{peters18-elmo}, a more light-weighted pre-trained model, to show the effect. For the pre-training tasks, we follow the same language model task of ELMo.

 We have three pre-trained models, including:
 
 
 (1) the officially released ELMo;
 
 (2) pre-trained model by initializing with random parameters and then performing meta-learning based pre-training;
 
 (3) pre-trained model by first initializing with official ELMo parameteres and then performing meta-learning based pre-training. 
 

We use SNLI~\cite{snli:emnlp2015} as the downstream task here for the former three ELMo variations. The experimental results are shown in Table~\ref{tab:elmo_snli}. 

\begin{table}[]
    \centering
    \begin{tabular}{l|c}
    \toprule
        Pre-trained Models & Accuracy  \\
    \midrule
        (1) Official ELMo  & 88.0\% \\
        \tabincell{l}{(2) Pre-trained ELMo with\\ random initialization} & 88.3\% \\
        \tabincell{l}{(3) Pre-trained ELMo with \\official initialization}  & 88.5\% \\
    \bottomrule
    \end{tabular}
    \caption{Fine-tuning results on SNLI with different pre-trained models.}
    \label{tab:elmo_snli}
\end{table}
Comparison between models (1) and (2) shows our approach has the ability to learn a better pre-trained model from scratch based on ELMo. Comparison between models (2) and (3) indicates the setting with a good pre-training starting point will obtain better results on downstream tasks.

\subsection{Analysis of Fine-tuning Initializations}
Experimental results in Table~\ref{unsupervised_result} show that our algorithm can obtain better results than official BERT when model converges. 
We go one step further to investigate
whether our model has a better initialization at the beginning of the fine-tuning phase.
We fine-tune our model on three datasets: SST-2, SST-5 and QDC for 4 epochs. 

Figure~\ref{fig:curve} shows the results.
We can observe that the pre-trained models with meta train step $k \geq 1$ can obtain better results than BERT at earlier epochs (e.g. epoch 1), which indicates that our learning algorithm can actually learn a better initialization for downstream tasks.

\section{Conclusion}
We introduce a learning algorithm which regards the pre-training of text representations as model-agnostic meta-learning.
We test our approach with multiple model architectures and multiple pre-training tasks. 
Results demonstrate the effectiveness of our approach. %

\bibliography{acl2020}
\bibliographystyle{acl_natbib}

\end{document}